\documentclass[conference]{IEEEtran}
\IEEEoverridecommandlockouts
\usepackage{amsmath,amssymb,amsfonts}
\usepackage{algorithmic}
\usepackage{graphicx}
\usepackage{textcomp}
\usepackage{xcolor}
\usepackage{pifont}
\def\BibTeX{{\rm B\kern-.05em{\sc i\kern-.025em b}\kern-.08em
    T\kern-.1667em\lower.7ex\hbox{E}\kern-.125emX}}

\usepackage[giveninits=true]{biblatex} 

\makeatletter
\newcommand{\linebreakand}{%
  \end{@IEEEauthorhalign}
  \hfill\mbox{}\par
  \mbox{}\hfill\begin{@IEEEauthorhalign}
}
\makeatother

\begin{document}

\title{Deriva-ML: A Continuous FAIRness Approach to Reproducible Machine Learning Models \\
\thanks{The work reported on in this paper was funded by an AI4Health collaborative grant from the University of Southern California.}
}

\author{\IEEEauthorblockN{Zhiwei Li}
\IEEEauthorblockA{\textit{Information Sciences Institute} \\
\textit{Viterbi School of Engineering}\\
\textit{University of Southern California}\\
Marina del Rey, USA \\
0009-0003-2848-7711}
\and
\IEEEauthorblockN{Carl Kesselman}
\IEEEauthorblockA{\textit{Information Sciences Institute} \\
\textit{Viterbi School of Engineering}\\
\textit{University of Southern California}\\
Marina del Rey, USA \\
0000-0003-0917-1562}
\and
\IEEEauthorblockN{Mike D'Arcy}
\IEEEauthorblockA{\textit{Information Sciences Institute} \\
\textit{Viterbi School of Engineering}\\
\textit{University of Southern California}\\
Marina del Rey, USA \\
0000-0003-2280-917X}
\linebreakand
\IEEEauthorblockN{Michael Pazzani}
\IEEEauthorblockA{\textit{Information Sciences Institute} \\
\textit{Viterbi School of Engineering}\\
\textit{University of Southern California}\\
Marina del Rey, USA \\
0000-0002-4240-7349}
\and
\IEEEauthorblockN{Benjamin Yizing Xu}
\IEEEauthorblockA{\textit{Department of Ophthalmology} \\
\textit{Keck School of Medicine}\\
\textit{University of Southern California}\\
Los Angeles, USA \\
0000-0003-1573-988X}
}

\maketitle


\begin{abstract}
Increasingly, artificial intelligence (AI) and machine learning (ML) are used in eScience applications~\cite{choudhary2023ai}.
While these approaches have great potential, the literature has shown that ML-based approaches frequently suffer from results that are either incorrect or unreproducible due to mismanagement or misuse of data used for training and validating the models~\cite{hanson2023garbage,kapoor2023leakage}.
Recognition of the necessity of high-quality data for correct ML results has led to \textit{data-centric} ML approaches that shift the central focus from model development to creation of high-quality data sets to train and validate the models~\cite{10.1145/3571724,10.1145/3411764.3445518}.
However, there are limited tools and methods available for data-centric approaches to explore and evaluate ML solutions for eScience problems which often require collaborative multidisciplinary teams working with models and data that will rapidly evolve as an investigation unfolds~\cite{10254870}.
In this paper, we show how data management tools based on the principle that \textit{all} of the data for ML should be findable, accessible, interoperable and reusable (i.e. FAIR~\cite{wilkinson2016fair}) can significantly improve the quality of data that is used for ML applications. 
When combined with best practices that apply these tools to the entire life cycle of an ML-based eScience investigation, we can significantly improve the ability of an eScience team to create correct and reproducible ML solutions.
We propose an architecture and implementation of such tools and demonstrate through two use cases how they can be used to improve ML-based eScience investigations.
\end{abstract}


\section{Introduction} 
Artificial intelligence (AI) and machine learning (ML)-based methods are increasingly important tools for diverse classes of eScience applications~\cite{choudhary2023ai}.
However, recent studies have shown that ML-based methods often suffer from significant issues related to correctness and reproducibility~\cite{kapoor2023leakage}. 
The effective use of ML-based methods is further complicated by the communication and coordination challenges that result from the collaborative, multidisciplinary nature of science teams, which may include domain scientists from multiple disciplines, data producers, research software engineers, ML engineers, etc.

High-quality data management and curation are essential for reproducible and correct ML-based science~\cite{hanson2023garbage}.
Data errors are often introduced early in the ML design process ``cascade,'' causing increasing difficulties as the process proceeds~\cite{10.1145/3411764.3445518}. 
Considering these factors leads one to ask: 1) What data should be curated? 2) When should it be curated? And, most importantly, 3) How should it be curated?
In previous work, we have explored these questions in a variety of settings, and have concluded that curation should apply to 1) all data, 2) all the time, and 3) via the creation of a \textit{data-centric socio-technical ecosystem}~\cite{dempsey2022sharing}. A data-centric infrastructure is one where the primary focus is on data and the organization of high-quality data collections rather than programs, processes, and models. A socio-technical ecosystem considers the interaction of social and technical aspects of systems, emphasizing the interactions between people, technology, and organizational structures. Following a data-centric rather than process-centric approach promotes the importance of data to correct ML; curating all the data all the time addresses the issue of data cascades and enhances transparency and reproducibility. Creating curated data within the context of a socio-technical framework enhances the ability to communicate across the perspectives and approaches of diverse team members, further decreasing the potential for error-inducing misunderstandings.

In this paper we explore these questions further in the context of ML-driven eScience application domains. The contributions of this paper are:
\begin{enumerate}
    \item We present a cloud-hosted platform for creating a collaborative social-technical ecosystem for ML development of correct and reproducible eScience applications. 
    \item We develop best practices for ML-based collaboration in eScience based on this ecosystem.
    \item We evaluate the best practices with its application in two real-world use cases. 
\end{enumerate}


\section{ML in collaborative eScience Environments\label{section-background}}
The overall process for developing ML-based eScience applications has a common structure~\cite{8804457}, independent of its application domain, which we illustrate with a representative example, the EyeAI project, and will explore in more detail in Section~\ref{section-usecase}.

In our example, a small team of around 15 members, comprising ophthalmologists, computer scientists, medical students, and clinicians, forms to develop an explainable ML algorithm that can determine the likelihood that an individual has Glaucoma, the leading cause of permanent vision loss worldwide. 
The standard approach to making such predictions is to photograph the retina (a \textit{fundus photograph}) and examine the relative size of the optic nerve and surrounding area (the cup-to-disk ratio).
Ratios above a specific threshold are indicative of Glaucoma.
Fundus photographs are readily available from an existing clinical diabetic retinopathy screening program, as periodic fundus imaging is part of the standard of care for diabetes patients.

The ML specialists in the team believe that the model for diagnosis will perform better if only the area around the optic nerve is used, rather than using the entire fundus image.
A nineteen-layer convolutional neural network (VGG19) is chosen to identify a bounding box that includes the optic nerve.
The fundus images are partitioned into training, validation, and test sets, the model is configured with a set of hyper-parameters using the training and validation datasets and evaluated with the test dataset. After fine tuning the model, the results are reviewed by the domain specialists.

Once the features, i.e. bounding box, have been extracted, the next step is to develop a model to diagnose which patients might have Glaucoma from that feature.
ML and clinical specialists discuss the interpretation of the raw data: whether labels are per examination, per patient, and per eye; if both eyes should be included; if low-quality images should be included or rejected; and what other clinical data, if any, should be included.
Training, evaluation, and test datasets are then created by selecting a subset of patients such that the datasets have relatively fixed ratios of glaucoma and non-glaucoma subjects. Note that care must be taken to ensure that there is no bias in the resulting data.
As with the bounding box algorithm, a model specialized in image classification is trained on the training and validation dataset. 
The ML team members again use VGG19 for this purpose, and the process of hyper-parameter selection, fine-tuning, training, validation, and testing is repeated on the datasets of patients, and the ophthalmologists evaluate the results.  
At that point, the team may decide whether the performance is satisfactory, or if not, the team may add new features, add additional data (such as clinical data elements), or try alternative modeling approaches to achieve their collective goals. 
Based on an analysis of ethnic cohorts, it is discovered that the ML algorithm under-performs for specific ethnic group members. To address this issue, additional data is obtained from the clinical partner; new training, validation, and test datasets are created that include criteria of ethnicity and diagnosis; and the entire training and evaluation process for the diagnostic model is repeated.

In practice, unfortunately, working methods are all too ad-hoc. Data dumps in the form of spreadsheets and directories full of images are transferred to the research team from the clinical partner. These are generally placed in a shared cloud-hosted (e.g. Google, Dropbox) drive. 
One of the ML engineers then writes one-off code to rename some of the files with provided diagnosis labels (likely-glaucoma, no-glaucoma). 
Another ML engineer subsets data into training/testing/validation datasets on a personal laptop and uses them to develop the bounding box identification model, generating a new set of images that must be partitioned into datasets for the diagnosis model.
The data is manipulated offline by uploading and downloading from Google Drive using one-off tools or processes, often saving the intermediate results as tabular data in a spreadsheet without creating a data dictionary and column headings whose meaning may be ambiguous.  
Experiment work is spread across researchers' laptops, whereas shared scripts and environments are located on GitHub, in Google Colaboratory (Colab), or in Virtual Machines, and the corresponding data may be distributed across each of these environments.
Depending on the task and the user's experience level, a wide variety of tools and computational environments are used, including Jupyter notebooks and Python scripts using libraries such as PyTorch, Excel spreadsheets, R programs, and special purpose tools.
Furthermore, collaboration requires extra effort to keep everyone updated on the latest data or model, and it is difficult for team members to review or reproduce the results of any one step or to interpret the end results of the model development in the context of all the data that went into producing that result.

The negative consequences of this current practice are pervasive in every stage of the project. 
Misunderstandings of the data; inconsistent data assignment into training, validation, and testing sets; and outdated datasets after feature engineering lead to data leakage and errors in the ML application. 
Offline scripts and manual manipulation make it hard to source the errors that happen upstream. 
Before applying our approach to the use case described above, we found hundreds of errors in the initial data (duplicated data, images without metadata, metadata without images, improperly labeled data); significant misunderstanding in interpreting the data (diagnosis per patient, per observation or per eye); lack of clarity in exactly what datasets were being used for each workflow step; and difficulty communicating and interpreting the results. 
All the problems mentioned above need data expertise, yet the team members typically lack it. 
In our experience, these practices are widespread and typical.
   
\section{Requirements and Approach \label{section-requirements}} 

An analysis of various collaborative eScience ML development projects, such as the one described above, leads us to propose a set of four requirements that should be met by any solution to the pressing issue of correctness and reproducibility in ML models for eScience. A solution should be:

\textbf{R1: Data-centric}. Well-trained ML models applied to bad data will produce bad results. 
While correct model code is important, we assert that the primary artifact of interest in ML-based eScience needs to be the data, rather than the algorithm and program. Hence, we should take a data-centric rather than process-centric approach~\cite{10.1145/3571724, schuler2015data}.
    
\textbf{R2: Comprehensive.} The challenge of data cascades implies that data management solutions should be applied to all data consumed and produced throughout the ML model development life cycle. 
    
\textbf{R3: Adaptive.} Any solutions need to be able to adapt to a variety of application domains, modeling approaches, and data types. 
In addition, eScience applications are science~\cite{kesselman2023let}, which by its nature will require that the solution evolve over the lifetime of an ML design experiment as new understanding is gained, the problem space is refined, and new approaches are explored.
    
\textbf{R4: Socio-technical.} A socio-technical system is one in which the interactions between people, community, software, and hardware are considered holistically~\cite{baxter2011socio}. 
Creating new ML-based eScience is a collaborative, iterative process that involves interactions both between team members and with the underlying computational environment. Solutions must streamline these frequent interactions (i.e., R2) even when team members come from different disciplines. 
 

\subsection{Approach \label{section-approach}}

To address the above requirements, we adopt an approach we call \textit{Continuous FAIRness}~\cite{dempsey2022sharing}.
Within the data-sharing community, it has become widely accepted that data in a repository will be more effectively used if that data is Findable (F), Accessible (A), Interoperable (I), and Reusable (R). Collectively, these have become known as the FAIR principles~\cite{wilkinson2016fair}. 

We have previously noted~\cite{dempsey2022sharing} that FAIR properties are advantageous not only for data in repositories but for all data generated during a scientific investigation.
We observed that ensuring that all data are FAIR can not only streamline the deposition of data into a repository of record but also accelerate the rate of discovery, by reducing the potential for error and enhancing the ability of a collaborative effort to communicate as the investigation progresses.
We use the term continuous FAIRness to refer to processes that adopt this approach.

Adopting continuous FAIRness as the foundation of our approach addresses all four core requirements. 
Continuous FAIRness implies that all data is FAIR all the time (R1, R2). 
FAIR data must have accurate and up-to-date metadata descriptions and be rendered in interoperable formats, facilitating communication across team members (R3, R4). 

Finally, a consequence of broadly adopting continuous FAIRness as the underlying approach is that support must be provided to ensure FAIR properties are seamlessly applied to every element of data created while developing an ML solution. Supporting this perspective encourages collaboration and interaction among team members around shared data. 
Even if the focus of a particular project is on developing a new model, the execution and evaluation of that model are described in terms of the data consumed and produced. 
Consequently, continuous FAIRness implies that data are the central artifact managed by our approach, aligning with the principles of Data-centric. 
\section{Architecture \label{section-architecture}} 
\begin{figure*}[t]
    \centering
    \includegraphics[width=1\textwidth]{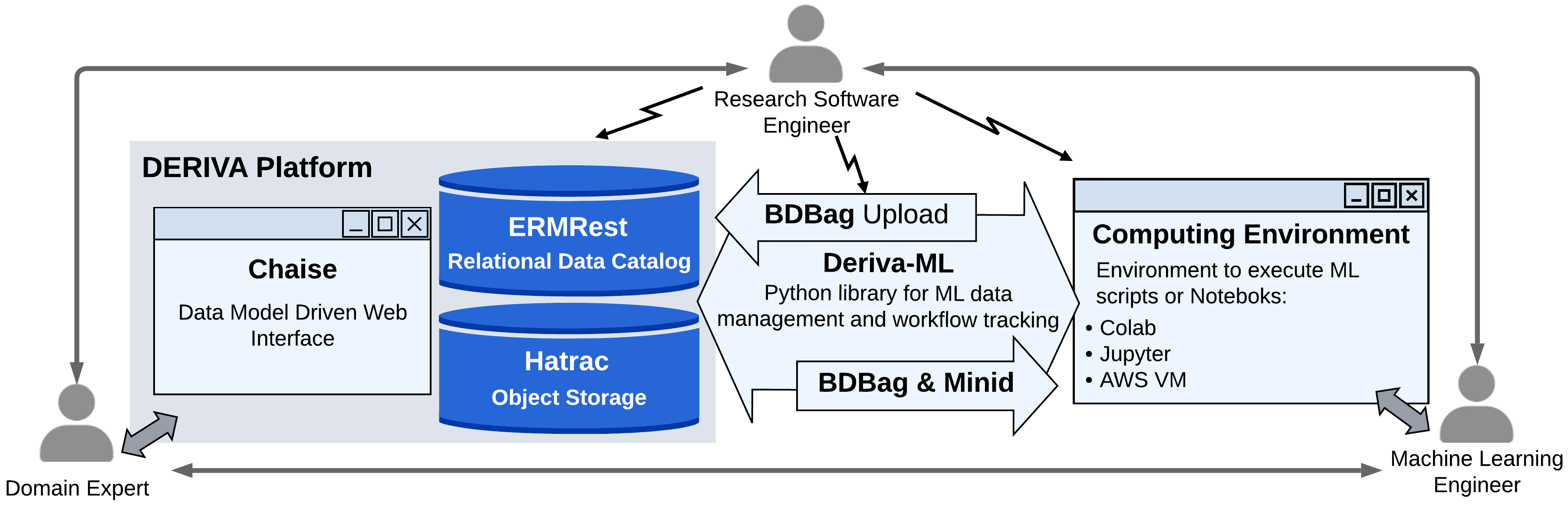}
    \caption{The Deriva-ML Architecture has three main components. The Deriva platform for scientific asset management is responsible for data discovery via a metadata catalog service (ERMRest), a versioned object store for storing all data assets (Hatrac), and a model-driven adaptive user interface (Chaise).
    Big Data Bags (BDBag) and a Python interface library, Deriva-ML, support data migration, and map lower-level Deriva operations into higher-level interfaces that more closely align with the ML development process. 
    We also support various on-demand computing environments used for shared ML development. 
    The key roles are the research software engineers who build the system, domain experts who curate the data through the Deriva platform, and machine learning engineers who develop ML models in computing environments.}
    \label{fig-architecture}
\end{figure*}

Fig.~\ref{fig-architecture} shows an overview of our system architecture for creating a collaborative, socio-technical ecosystem for conducting continuously FAIR, ML-based eScience investigations.
On the technical side of the architecture, we build on the Deriva scientific asset management platform (described in the next section) and augment it with additional services and Deriva-ML, a programming library designed to support the general ML development process.

On the social side of the system, we consider the expertise and interactions between three different roles. The domain expert understands the problem being solved and how to interpret the data and the results.  
While they may have some familiarity with ML methods, they are not an expert. 
They may have programming skills in Python notebooks or R but are not expert coders. 
On the other hand, machine learning engineers have experience developing machine learning methods using tools such as PyTorch or TensorFlow; however, they typically do not have deep domain expertise. 
They may have sufficient programming skills to develop domain-specific extensions to the Deriva-ML class libraries described below but limited expertise in data management.
Finally, we have the role of the research software engineer (RSE)~\cite{rse}, who understands data management and data modeling and has the skills to create significant extensions of the underlying Deriva-ML library.
Next, we will highlight some key components and their implementation.


\subsection{The Deriva Platform}
Deriva~\cite{8109125} is a collaborative scientific asset management platform. 
The platform is designed to support collaboration through the entire scientific data life cycle, including the initial experiment design, prototype, production data acquisition, ad hoc and routine analyses, and publication.
The core principles underlying the design of Deriva are:
\begin{itemize}
\item loosely coupled web services architecture with well-defined public interfaces for every component, 
\item use of Entity Relationship Models (ERM) that leverage standardized vocabularies, with adaptive components that can automatically respond to evolving ERMs,
\item model-driven user interfaces to enable navigation and discovery as the data model evolves, 
\item data-oriented protocols where distributed components coordinate complex activities via data state changes.
\end{itemize}

Deriva was designed to support the creation of socio-technical platforms for data-driven collaboration, with FAIR principals being an essential enabler of such collaboration. 
 

\subsection{Deriva-ML Metadata Catalog Design}

A critical aspect of FAIR data management is having accurate and relevant metadata for all data objects associated with the ML development.
Within Deriva-ML, this metadata is described explicitly using an ERM and implemented via the ERMRest catalog service, which provides a well-defined RESTful web services interface that maps ERM functions onto relational database management software (Postgres).

In structuring the ERM, we observe that the data from the domain area will vary from application to application, but the data generated from the overall ML development process, such as partitioning data into training and validation subsets,  specifying model parameters, running training and validation steps, collection runtime environment logs, etc., will be common across different application domains, as described in Section~\ref{section-background} and~\cite{8804457}.

\begin{figure}[htbp]
    \centering
    \includegraphics[width=0.45\textwidth]{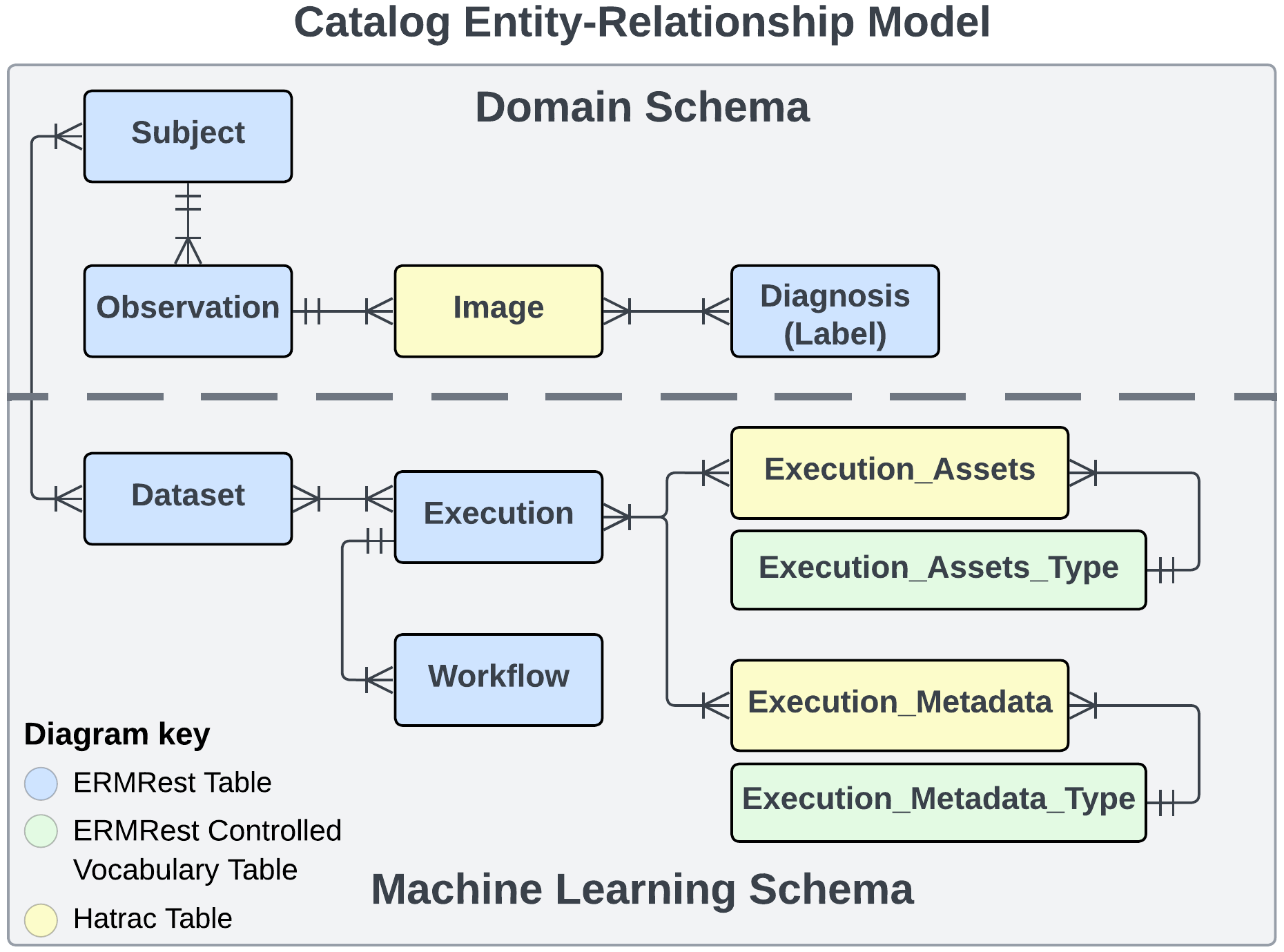}
    \caption{
        \label{fig-ERD}
    The metadata catalog is structured into two interconnected components. The upper component is tailored to a specific domain. In the upper part of the figure, we show the instantiation used for the use case in Section~\ref{section-background}. The lower part of the figure shows the data model for the reusable part of the catalog that models the overall ML data objects produced by the ML development process. The model consists of descriptions of the data element (in blue), data assets (in yellow), and controlled vocabulary (in green).
    }
\end{figure}

To maintain the adaptability for new domain areas while being able to reuse significant elements of the ERM,  we segment the ERM into two interlinked components, shown in Fig.~\ref{fig-ERD}.  One component represents the shared, reusable concepts that come from the overall ML development process, and the other captures the details of the application domain.
ERMRest supports placing ERM elements into named schemas, and we utilize that feature to implement catalog segmentation.

In this design, the domain part of the metadata model will be specialized to the domain and evolve over time, where the ML components of the catalog are mostly reused and remain stable over the lifetime of the model development.
When initiating a new ML application, the project RSE or ML-Dev must only define the domain-specific concepts and link to the more detailed process schema.
In practice, the initial version of the domain-specific part of the schema may consist of only a few concepts and attributes, and Deriva evolution features may enrich this model over time. 
This factoring allows a new application to be deployed in Deriva-ML rapidly by team members with limited experience in data modeling.

\subsubsection{Domain schema}
The domain schema includes the data collected or generated by domain-specific experiments or systems, and the ER model of the domain catalog will vary according to the application and domain knowledge.
In the upper part of Fig.~\ref{fig-ERD}, we illustrate the domain catalog with the use case mentioned in Section~\ref{section-background}, with multiple subjects, each having multiple observations consisting of one or more images and associated diagnoses. 

\subsubsection{ML schema}
Each entity in the ML schema is designed to capture details of the ML development process. 
There is only one connection between the ML schema to the initial domain schema, helping to maintain the flexibility of the ML schema in serving any domain application.
The use of persistent unique identifiers (PID) for all data records, rich metadata, and controlled vocabularies contributes to the reproducibility of ML experiments. 

The main entities in the ML catalog are as follows:

A \textbf{Dataset} represents a data collection, such as aggregation identified for training, validation, and testing purposes. Each dataset uses checksums to validate its contents and is assigned a persistent identifier and a set of dataset-specific metadata to facilitate findability and reuse.

A \textbf{Workflow} represents a specific sequence of computational steps or human interactions. A workflow may reference scripts or notebooks or any general workflows of end-to-end ML experiments, from data loading to data pre-processing, ML model development, and persisting results to the catalog. 
Workflow records have a PID and are described using terms from a controlled vocabulary.

An \textbf{Execution} is an instance of a workflow that a user instantiates at a specific time. It keeps track of the user, workflow, status, datasets, input and output files, and duration associated with the execution. 

An \textbf{Execution Asset} is an output (e.g., file) that results from the execution of a workflow.
Execution assets can include ML models as exported by libraries such as TensorFlow, hyper-parameter files used to configure a model, or output prediction results generated by applying the model to a dataset. 
Every execution asset has a PID and is described with terms from a controlled vocabulary.
    
An \textbf{Execution Metadata} is an asset entity for saving metadata files referencing a given execution. 
Common execution metadata are configuration files and runtime environment logs. 
This entity can also reference other self-defined metadata in arbitrary file formats. 

\subsection{FAIR Data Exchange}
Datasets shared among researchers can be large (many terabytes or even petabytes) and may comprise large numbers of individual files. Often, ML developers will ``stage" (create copies of) data files on storage systems that are ``close'' to the execution platform being used, a time-consuming and potentially expensive operation.
An unfortunate consequence of the resulting copies is that it becomes difficult to keep track of exactly which dataset is being used for a specific execution. Copies of a dataset may be modified (making it difficult to reproduce results), missing or added data files may go undetected. In addition, team members may unknowingly make copies of datasets that are already located on their desired computing platform, wasting both time and storage resources.

While these problems may be mitigated by adopting operational conventions within a project, providing additional software support within the underlying data management framework is a more desirable solution.
To ensure the efficiency and FAIRness of large dataset sharing, we leverage the \textbf{BagIt File Packaging Format}~\cite{bagit} as described in RFC 8493 to describe the data collection and use \textbf{Minid} as the persistent identifier~\cite{7840618}.
\textbf{BagIt} is a set of hierarchical file system conventions designed to support disk-based storage and network transfer of arbitrary digital content. A ``bag" consists of a ``payload" (the arbitrary content) and ``tags," which are metadata files intended to document the storage and transfer of the bag. A BagIt \textit{bag} provides a lightweight, reproducible description of an entire dataset's contents, suitable for reliable storage and transfer through the use of metadata file manifests containing checksum values for every file in the payload. We use BagIt as a mechanism for defining a dataset and its contents by enumerating its elements, regardless of their location, and to facilitate the assembly, reliable sharing, and analysis of such datasets. 

Our integration of BagIt into Deriva-ML uses the BDBag Python software package~\cite{bdbag}. 
In addition to providing a complete reference implementation of the BagIt specification, BDBag supports \textit{bag idempotency}, or reproducible bags. 
A reproducible bag is a bag that has content-equivalence (in both payload and metadata) to another bag created at a different time with the same content, structure, bagging tool, and profile. 
With bag idempotency, two separately created bags (or bag archive files) with content-equivalence will hash equally, whether the hash is calculated on the bytes of a bag archive file or calculated on the equivalently ordered set of individual file hashes of the bag's contents. 
We leverage bag idempotency to effectively enable reproducible caching of files on a storage system co-located on a computing environment used for ML model development. 
In practice, this means that researchers do not need to wait for an additional data download if a dataset with the same content has already been cached in the environment. 
This functionality dramatically improves the efficiency of data retrieval in ML workflows.

A \textbf{Minid} is a \textit{minimal viable identifier}, which is designed to be actionable, disposable, resolvable, and persistent. It includes extensible metadata and checksums to ensure data integrity and FAIRness. Minid, as a lightweight identifier, is easy to associate with a dataset and share among researchers.
By assigning a Minid to a BDBag when it is generated, we meet the FAIR criteria for permanent identifiers, and we facilitate reliable communication of datasets between team members, as a specific dataset can be unambiguously referred to by its Minid, regardless of its location.

\subsection{Controlled Resource Access}

Securing access to resources is a fundamental requirement of any collaborative environment. 
Our architecture comprises multiple resource components, all working in concert to facilitate eScience goals. 
Our implementation utilizes the Globus platform for much of this functionality, specifically Globus Auth and Globus Groups ~\cite{tuecke2016globus}. 
By leveraging Globus as the identity and access management (IAM) service for Deriva and JupyterHub, we can support a single set of user credentials for access to metadata, files, and compute resources. 
Furthermore, Globus Auth's role as an identity aggregator/broker greatly simplifies the on-boarding process for gaining access to our systems, as users can essentially ``bring their own" identities from any identity provider (IDP) supported by Globus Auth, e.g., Google, ORCID, federated IDPs (like \textit{inCommon}), or institutional (Campus) OIDC servers.  

Policy-based access control is another important aspect of our support of the roles and practice of collaboration. 
We can implement and evolve policy over the collaboration life cycle using Globus Auth, Globus Groups, and the fine-grained access control lists (ACLs) in the Deriva components (specifically in ERMRest and Hatrac). 
For example, our default configuration supports a ``self-curating'' policy that revolves around three central roles: \textit{reader}, \textit{writer}, and \textit{curator}. 
The self-curation policy model allows for a \textit{writer} user to own and curate self-created content, which can then be made selectively viewable, but not modifiable, to \textit{readers}. 
The role of \textit{curator} allows viewing, creating, and modifying all content.

This default policy allows for common use-case scenarios such as an ML student researcher/developer (a \textit{writer}) working iteratively on model development and being able to share partial results with a professor or supervisor (a \textit{curator}) without having to share intermediate results more broadly (i.e., to \textit{readers}). As the collaboration grows, policies can be further extended to support more sophisticated scenarios, such as a time-based data embargo or a consensus-based data review policy that only allows read access after specific conditions are met and approved by a set of curators.  

\subsection{The Deriva-ML Library}

Deriva-ML contains a Python library designed to support data migration between computing environments and data storage, automate ML process tracking, and encapsulate repetitive data tasks. 
Deriva-ML uses the Python class mechanism to provide generic functionality for manipulating data elements typically found in ML applications. 
Deriva-ML is designed to be used as a base class, which can be extended with domain-specific operations driven by the needs of the ML application. 
In designing Deriva-ML, our goal was to abstract out as much detail about the underlying representation as possible so that creating the derived class would be well within the skill set of a typical ML researcher. 
To reduce errors and streamline the creation of specialized classes based on Deriva-ML, the library follows modern Python coding standards, including extensive use of type hints, data classes, and embedded doc-strings.
 
\begin{figure}[htbp]
    \centering
    \includegraphics[width=0.4\textwidth]{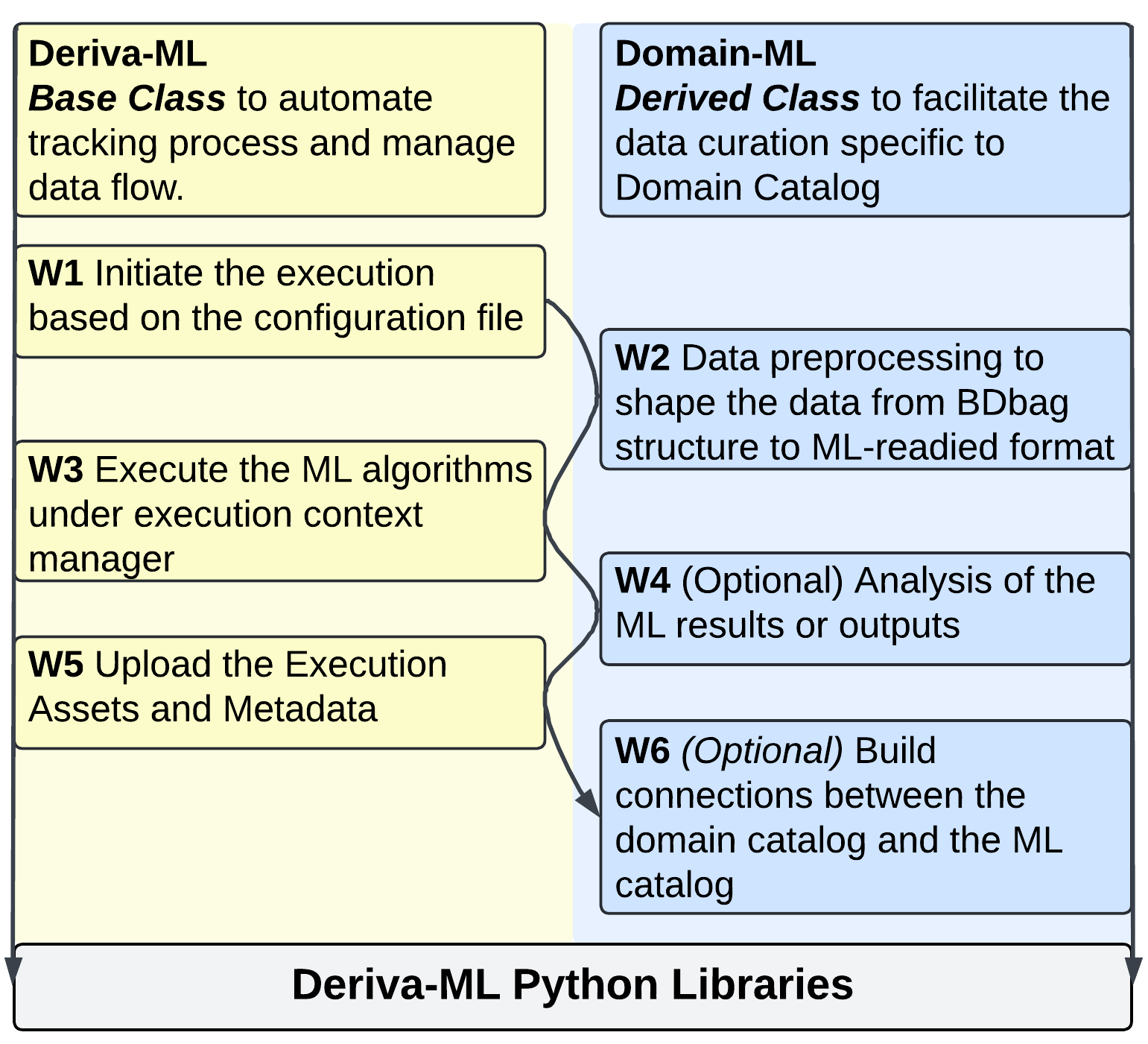}
    \caption{
        \label{fig-library}
    The main components and methods of Deriva-ML library. The left section is the base class, Deriva-ML, which contains three essential methods to support the ML process tracking that is general to various applications. The right section shows the derived class Domain-ML. It contains three types of methods customized to the application and domain data model for efficient data management. The components W1 through W6 facilitate the entire ML workflow, with W4 and W6 optional.}
\end{figure}

A core function of the Deriva-ML library is to provide an interface between the data-centric structure of Deriva-ML and the computational environment used for modeling and analysis. Before a modeling or analysis step, datasets must be transferred to the computing platform based on their identifier and structured in a local file system according to the specifics of the modeling or analysis step, then computationally processed, and finally, the resulting data characterized with controlled vocabulary terms and reintegrated into the Deriva-ML catalog.

To accommodate a wide variety of modeling techniques and execution environments, including notebooks, automated workflows, and manual creation, we need to capture the relationship between data and associated methods of generation in a general way.
We accomplish this with a simple provenance model which identifies a workflow with a series of steps and an execution instance that connects the workflow with a computing environment and input and output data and file assets. Every element of this model is provided with a globally unique persistent identifier: Minids in the case of data/file assets and GitHub URLs with a cryptographic hash for scripts and code.
Within Deriva-ML, the connection between an execution instance and a workflow is captured via a simple JSON configuration file, which uses the permanent identifiers of the data and code to specify the parameters of the execution.

Fig.~\ref{fig-library}) summarizes the essential steps in an ML workflow and reduces the friction at data migration.

The implementation of the methods in each class is as follows:
\subsubsection{Deriva-ML base class}
\begin{itemize}
    \item Execution initiate: Initialize a workflow via a structured configuration file. All inputs, such as the dataset bags and models in the Execution Assets table, will be downloaded or cached into the computing environment. 
    \item ML execution context manager: A Python context manager that initiates an execution. It facilitates error logging to the catalog and manages the resources and memory when ML experiments fail.
    \item Execution upload: At the end of an execution, this method will upload all the newly generated files from a well-known location to the catalog. Files that are computational products like new features, predictions, and models are uploaded to the Execution Assets table. Metadata files will be uploaded to the Execution Metadata table, which by default includes the execution configuration file and a runtime environment log. 
\end{itemize}

\subsubsection{Domain derived class}
\begin{itemize}
    \item Data pre-processing: Transforms the data from bag structure to ML-readied format, involving data tables joining and filtering, file reorganization, etc.
    \item Data Analysis: Examples include using the predictions to calculate or plot the model evaluation metrics.
    \item Relationship building: These methods will build relationships between the Execution Asset table and the domain catalog. Examples are linking the new feature from the engineering process to the original features, connecting the prediction to the labels, etc.
\end{itemize}

\subsection{Computing Environment}
Researchers can access the Deriva catalog via the  Deriva-ML Python API in various environments, such as a local Jupyter Notebook environment, local script environment, and Google Colab.
Our production compute system comprises a pre-configured, multi-user Jupyter Hub environment with GPU resources and Deriva-ML. 
This environment allows for quick and fully tracked ML experiments to be performed. The shared data caching directory can further enhance the data migration and sharing efficiency among research teams. Additionally, the connection between the Jupyter Hub environment and GitHub seamlessly manages the workflow versioning and development of the Deriva-ML domain class.

\section{End-to-End best practice}

\subsection{Data Modeling}
A key aspect of our data-centric approach is to create and document an explicit data model, i.e. an ERM.
Collaborations between domain experts and data experts over a shared ERM facilitates communication and provide the foundation for metadata descriptions needed for reproducibility and sharing.
Our current implementation uses Lucidchart~\cite{lucid-chart}, a web-based diagram application that supports team collaboration and diagram versioning, to develop ERMs. With embedded links to each entity in the catalog, the ERM on Lucidchart serves as a map for navigating the whole project catalog.
As the scientific investigation progresses, the domain schema will evolve in response to the shift of problem space. 
Deriva mechanisms for model evolution are used to implement these changes~\cite{schuler2022managing}.

\subsection{Controlled Vocabulary Development}
Controlled vocabularies are lists of pre-defined data elements that describe terminologies in a central domain to ensure consistency in communication and data organization. 
The evolution of shared vocabulary is an essential aspect of collaboration~\cite{noy2004ontology}.
These vocabularies help to create a shared language across the research team, further improving communication and streamlining collaboration.
Controlled vocabulary terms are explicitly represented in the platform, and the user interface automatically adapts to include terms as potential values when characterizing data.
Elements in the controlled vocabulary can be widely agreed upon coding schemes or can be 
 local to the collaboration and defined by the team.

\subsection{Loading and Partitioning Data}

eScience data ingest tends to be infrequent and idiosyncratic, and standardized Extract/Translate/Load tools are not widely adopted. One-off data ingest and cleaning procedures using tools such as Python, R, and shell scripts are typical.
Deriva-ML standardizes these processes via a robust set of command line tools and Python APIs that support simultaneously uploading both data files from a local file system into the object store and associated file metadata to the catalog.

Partitioning data into well-curated subsets, such as training, testing, and validation sets, is an essential part of  ML-based projects. Rather than relying on directory structures or spreadsheets to capture datasets, the Deriva-ML data model explicitly represents a data collection that uniquely identifies all of the file assets, their contents, and all the metadata associated with the collection.
Further codified by Minid and BagIt, these datasets can be reliably transferred to a compute/analytic platform in a reproducible manner.

\subsection{ML Development}
The ML development process builds on  Deriva-ML to track all associated scripts, data, and metadata. The final products are workflows in the catalog for training, testing, and evaluation, with instantiated executions related to all of the precise inputs, results, and metadata necessary for reproduction.

To better compartmentalize the code according to role, we create two separate GitHub repositories for collaborative code development: the first contains Python modules that implement the ML models and the interface library derived from Deriva-ML, and the second contains notebooks, scripts, and auxiliary programs that combine with Deriva-ML and it's domain-specific derived classes to form an end-to-end analysis pipeline.  This separation of repositories also allows for the more strictly versioned library code and modules to be pre-installed into the computing environment, whereas workflow scripts can be run in a more ad-hoc, experimental or iterative manner. 

To reflect the main aspect of the ML software engineering process, we  follow the workflow structure shown in Fig: \ref{fig-library}, which divides each ML workflow into six (W1-W6) subsections. Each subsection is an independent module to be reused and debugged.


\subsection{End-to-end Best Practice}
\begin{figure}[htbp]
    \centering
    \includegraphics[width=0.45\textwidth]{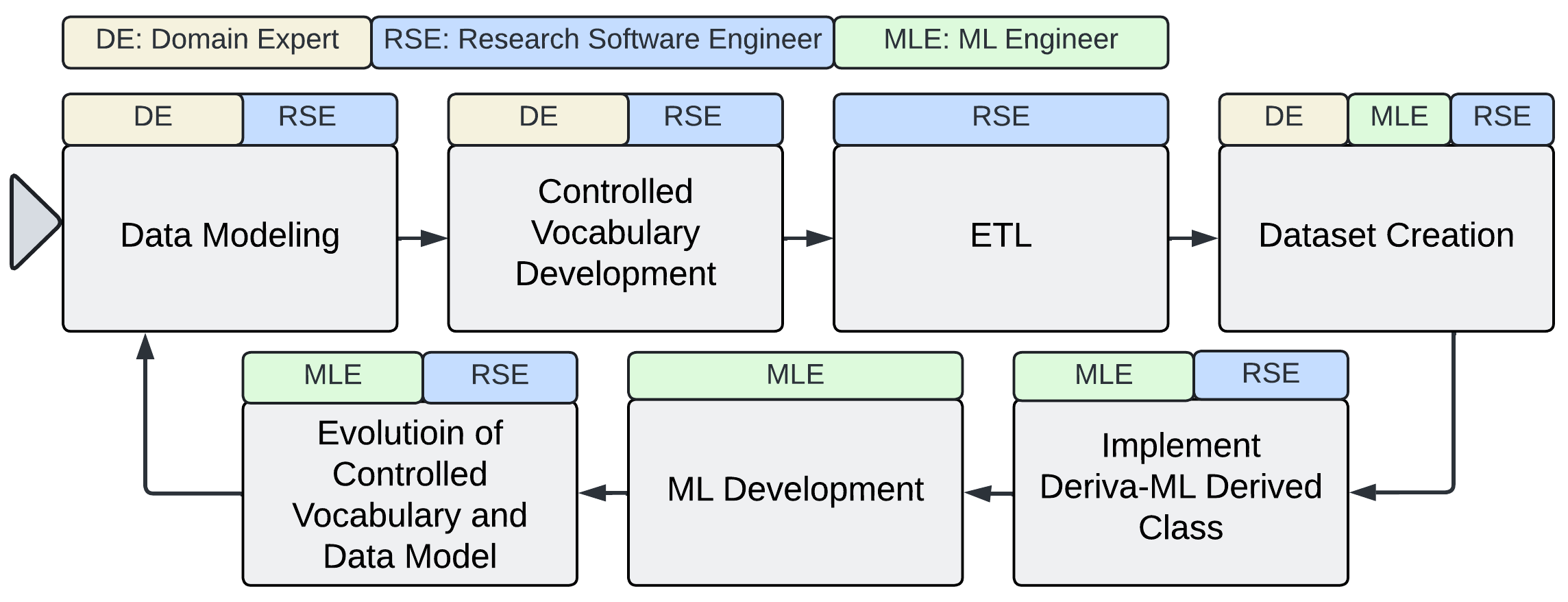}
    \caption{\label{fig-workflow}
    End-to-end data-centric best practice involving seven steps. The workflow is cyclical, with the last step (Evolution of controlled vocabulary and data model) looping back to the first step (Data Modeling), reflecting the iterative process. The essential roles involved are shown at the top of each box, which includes Domain Experts, Research Engineers, and ML Engineers.
    }
\end{figure}

We summarize our best practice as seven actionable steps in Fig:\ref{fig-workflow}. It starts with Data Modeling and ends with the Evolution of the Data Model and Controlled Vocabulary, which then loops back to the first step to form a cycle. Our proposed best practice emphasizes the \textit{data} as a collaborative foundation continually iterated by domain experts, ML engineers, and research software engineers to achieve high-quality, reproducible ML-based eScience projects.

\section{Use Case\label{section-usecase}} 
This section presents two use cases to illustrate how an eScience team can use this data-centric architecture and the ``Best Practice" to develop reproducible ML products.
\subsection{Use case 1: EyeAI - Glaucoma Detection}

\subsubsection{FAIR data catalog and library setup}
The background of this project has been introduced in section~\ref{section-background}. 
The team modeled the data by collaborating on the shared Lucidchart as shown in Fig: \ref{fig-ERD}. 
The controlled vocabulary is also developed for the image side (Left/Right/Unknown), diagnosis labels (Referable Glaucoma/No Glaucoma/Unknown), and demographic information like gender, ethnicity, etc. 
While developing the ERD and controlled vocabulary, the issue of where diagnosis should be placed (eye, observation, subject) was highlighted and identified, and using the ERD, a collective decision was reached across the team.

The EyeAI-ML derived class is instantiated from Deriva-ML with the EyeAI catalog's domain and ML schema.  It features four primary methods to facilitate the workflow: two for data pre-processing, including image filtering by viewing angle and cropping via a bounding box for the optic nerve area, and two for data analysis, focusing on extracting diagnoses from images and plotting the ROC curve for quick performance evaluation.
Meanwhile, aligning with the ML development plan, two sub-modules are created to develop models to identify the bounding box and predict the diagnosis. 

In discussion across the team, it was decided that the training, validation, and test dataset should contain 20\%, 60\%, and 20\% of the total subjects, respectively, with the same portion of diagnosis labels for each class based on a subject-level diagnosis. 
It was also determined that additional labeling should be performed to determine the level of consensus across multiple ophthalmologists reviewing the images in a traditional manner.
To enable this, smaller datasets were created from the three standard ML datasets for clinicians to provide their diagnosis to improve the label quality and evaluate the model performance.

Using the EyeAI-ML library, a simple customized reviewing tool was constructed, and a panel of five ophthalmologists was recruited to review the images.  The results were by associating additional diagnosis terms with each image and subject, resulting in a collection of diagnoses per subject.
The EyeAI-ML library was then used to retrieve the diagnosis into a Python Pandas DataFrame, and various scoring methods were used to compute a ``gold standard'' label for each data.

\subsubsection{ML development workflow}
To predict referable Glaucoma cases, the team mainly explored two ML models consecutively: the first is to identify the optic nerve from the image, and the second is to predict Glaucoma likelihood based on the optic nerve image.

The following is the workflow in a notebook to use a pre-trained model in the catalog to identify the bonding box and save the output bounding box in SVG in the catalog. The workflow matched the steps in Fig: \ref{fig-library}

\begin{description}
    \item[\textbf{W1}:] \verb|execution_init| Insert a new identifying bonding box workflow into the Workflow table. Create a new execution as the instance of the workflow. Load the ML model and dataset, including the raw images, into the computing environment. Create a new Execution Assets type: ``Image Annotation" for the output SVG files. 
    \item[\textbf{W2}:] \verb|filter_angle_2| Filter the angle-two images as only angle-two fundus images contain a clear and complete optic nerve.
    \item[\textbf{W3}:] \verb|with execution(exec_id) as exec| Run the ML algorithm with the context manager to raise the potential errors in the algorithm. The algorithm outputs bounding box SVGs into the compute environment.
    \item[\textbf{W4}:] No results analysis is involved in this workflow.
    \item[\textbf{W5}:]  \verb|execution_upload| Upload all the SVG files in the computing environment to the Execution Assets table.
    \item[\textbf{W6}:] \verb|insert_image_annotation| Build associations between the image and bounding boxes.
\end{description}

\begin{figure}[htbp]
    \centering
    \includegraphics[width=\columnwidth]{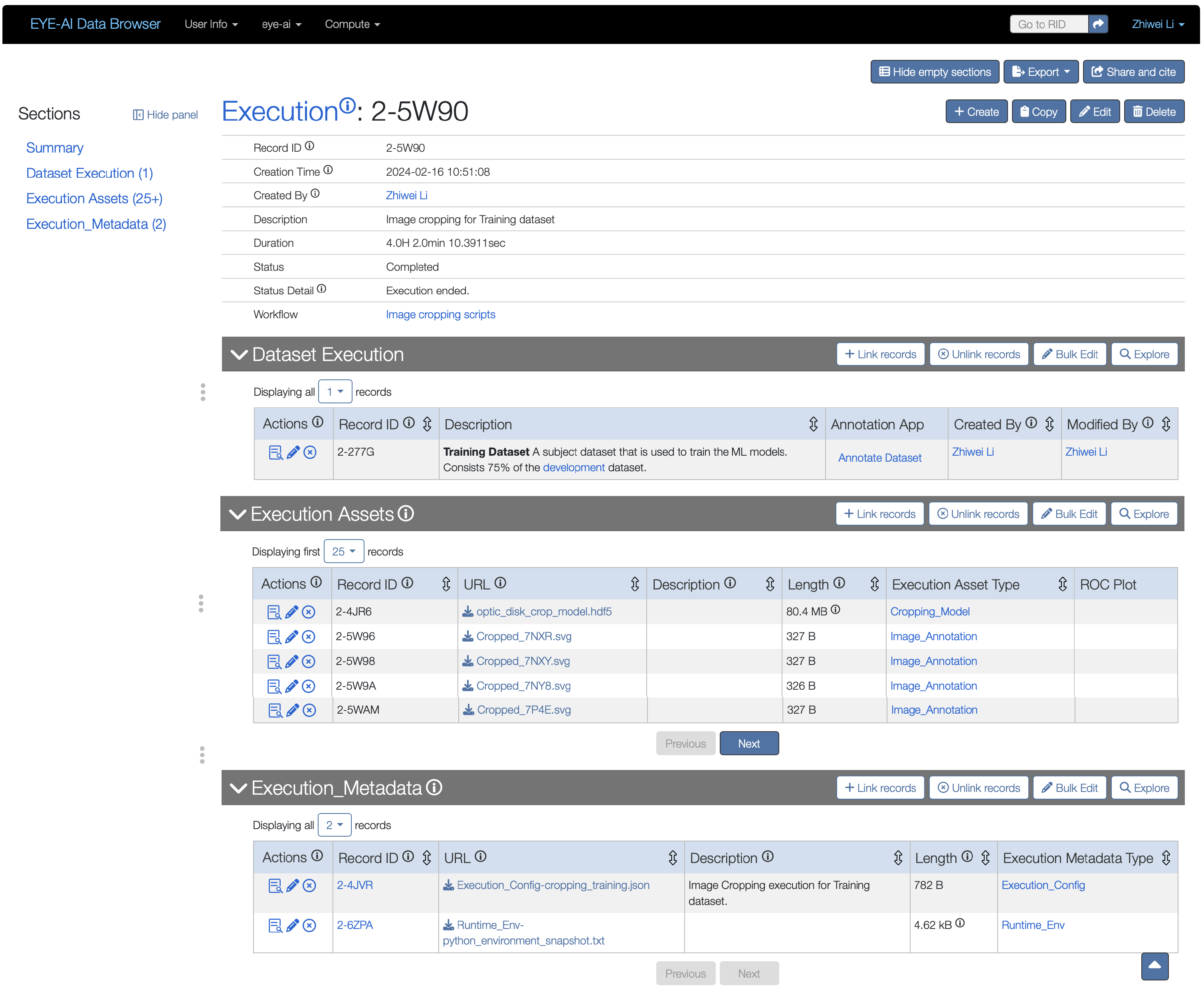}
    \caption{The detailed page of an Execution in the Chaise UI. It contains the workflow, datasets, execution assets, and execution metadata associated with the execution to ensure reproducibility.}
    \label{fig-execution-EyeAI}
\end{figure}

After the execution is completed, the whole process is tracked and written back to the catalog, as shown in Fig: \ref{fig-execution-EyeAI}. 
The execution and outputs can be easily and precisely reproduced using the metadata provided on this page.

In the subsequent phase, we developed the diagnosis model by adhering to the workflow in Fig: \ref{fig-library}.
A notable implementation of the EyeAI-ML library is creating cropped images during W2 and plotting the ROC curve at W4. 
Upon a detailed examination of prediction results, it was observed that the model was biased toward the Latino subjects. By tracking back to the training datasets, Latino subjects were found to be underrepresented in the dataset. 
To enhance the fairness of the model, additional data records were extracted from the clinical sites and went through the same ETL process to be included in the catalog.


\subsection{Use case 2: MusMorph Genotype Prediction}
\begin{figure}[htbp]
    \centering
    \includegraphics[width=0.45\textwidth]{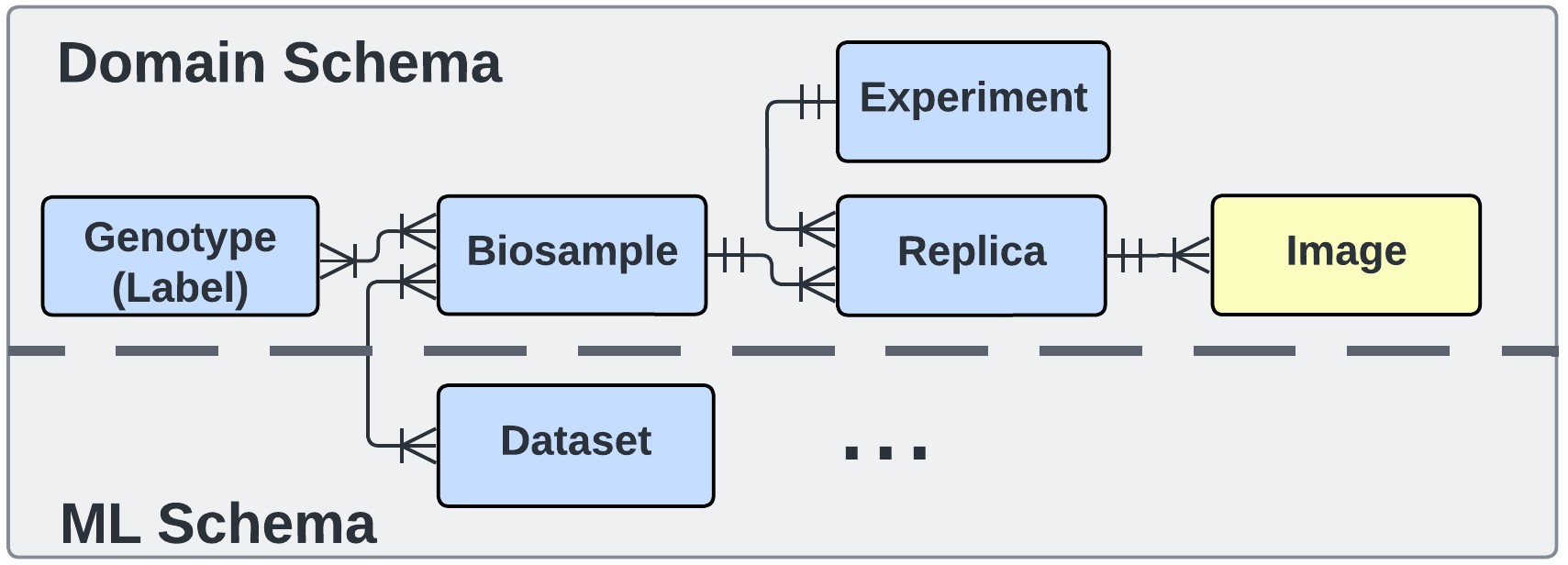}
    \caption{The metadata catalog for the MusMorph project. The upper part is tailored to collect the tomography data with the metadata from experiments in the laboratory. It includes the information on the bio-sample, its corresponding replica, experiments on the replica, and image files from the micro-CT scanner. The lower part of the figure models the ML data objects.}
    \label{fig-FB-ERD}
\end{figure}

MusMorph is a sub-project of FaceBase~\cite{schuler2022facebase} containing standardized mouse morphology data. The data provenance differs from the Glaucoma detection use case; instead of extracting the data from the existing system, the imaging data and metadata are collected through meticulously planned experiments, aligning with the Entity-Relationship Model depicted in the Fig.~\ref{fig-FB-ERD}. 

An ML model is developed to predict the genotype of the mouse bio-sample according to their skull micro-computed tomography (micro CT). The ML Catalog is the same as the standard design in Fig: \ref{fig-ERD} except the datasets are connected to the Biosample table. The training process follows the standard workflow mentioned in Fig.~\ref{fig-library}.

\begin{description}
    \item[\textbf{W1}:] \verb|execution_init| Insert training workflow and a new execution instance. Load the datasets. Create a new Execution Asset Type, ``Genotype Prediction."
    \item[\textbf{W2}:] \verb|load_images_labels| Summarize the genotype of the bio-sample from types to control and experiment group.
    \item[\textbf{W3}:] \verb|with execution(exec_id) as exec| Run the training algorithm to train the model and output a model object.
    \item[\textbf{W4}:] \verb|plot_roc| Plot the ROC curve of the model on the test dataset.
    \item[\textbf{W5}:] \verb|execution_upload| Upload the model objects and corresponding ROC plot to the catalog.
    \item[\textbf{W6}:] No connections are built between the domain and ML catalog in this training workflow.
\end{description}

The training process is also tracked in the Execution table, including the training workflow script, datasets, output models, and model performance visualization, ensuring the model is reproducible and supports the comparison of model performance.

\subsection{Evaluation}
In this section, we evaluate the implementations of each use case. 
We then employ the FAIR Metrics \cite{fairshake2024} to assess the FAIRness of the Deriva Catalog. We ticked all of the satisfied metrics for each use case in table~\ref{tab:evaluation-FM}.


\begin{table}[ht]
\centering
\caption{FAIR Metrics v.s. Use case}
\label{tab:evaluation-FM}
\begin{tabular}{|c|c|c|}
\hline
\textbf{FAIR Metrics} & \textbf{Glaucoma} & \textbf{MusMorph} \\
\hline
Globally unique identifier & \ding{51} & \ding{51}\\
Persistent identifier & \ding{51} & \ding{51}\\
Machine-readable metadata & \ding{51} & \ding{51}\\
Standardized metadata & \ding{51} & \ding{51} \\
Resource identifier in metadata & \ding{51} & \ding{51}\\
Resource discovery through web search & \ding{51} & \ding{51}\\
Open, Free, Standardized Access protocol & \ding{51} & \ding{51}\\
Protocol to access restricted content & \ding{51} & \ding{51}\\
Persistence of resource and metadata & \ding{51} & \ding{51}\\
Resource uses formal language & \ding{51} & \ding{51}\\
FAIR vocabulary & \ding{51} & \ding{51}\\
Linked &  & \\
Digital resource license & \ding{51} & \ding{51}\\
Metadata license & \ding{51} & \ding{51}\\
Provenance scheme & \ding{51} & \ding{51}\\
Certi. of compliance to comm. standard &\ding{51} & \ding{51}\\
\hline
\end{tabular}
\end{table}

\section{Related Work \label{section-related}} 

MLOps manages the entire life cycle of machine learning models, focusing on creating ``Production-ready ML products'' for organizations requiring frequent retraining and redeployment of models~\cite{kreuzberger2023machine}. 
Domain data management and evolution needs are not adequately addressed by most MLOps products, which adopt a model-centric or process-centric approach.

An overview of data-centric AI tools is provided in ~\cite{jakubik2024data}. In general, the tools described are not focused on domain-specific customization required for eScience and do not support the FAIR data practices desired for scientific data sharing.

There are similarities between our approach and that taken by WholeTale~\cite{BRINCKMAN2019854}, which integrates the data, code, provenance, and lineage to promote reproducibility. However, WholeTale is not data-centric and does not provide the detailed metadata support for FAIRness that Deriva-ML does.

\section{Conclusion and Future Work \label{section-conclusion}} 
This paper addresses the correctness and reproducibility problems in ML-based science. We proposed a data-centric architecture with the Deriva Platform and Deriva-ML by analyzing the requirements of a collaborative reproducible ML project. Within our use cases, we have demonstrated that members of the ML development team can readily adopt the tools, they facilitate communication, and they can help detect and correct data-oriented errors early in the development process.
While promising, our results are mostly anecdotal, and we plan to undertake a more systematic evaluation of the effectiveness of our approach. 



Based on the use cases and user feedback,  areas for future work, include more robust code versioning management, streamlining the Minid-to-data objects mapping, and simplifying the ML workflow configuration process.

\section*{Acknowledgment} 

We thank our collaborators who helped develop our use cases: Sreenidhi Iyengar Munimadugu, Aniket Kumar, Van Nguyen, Lauren Daskivich, Shriya Nagrath, Robert Schuler, and Maryam Ahmadi.

\printbibliography 

\end{document}